\documentclass[letterpaper, 10 pt, conference]{ieeeconf}  
\usepackage[T1]{fontenc}
\usepackage{graphicx}
\usepackage{cite}
\usepackage{amsmath,amssymb,amsfonts}
\usepackage{algorithmic}
\usepackage{array}
\usepackage{multirow}
\usepackage{textcomp}
\usepackage{booktabs}
\usepackage{bbding}
\usepackage{xcolor}
\usepackage{pifont}
\usepackage{url}
\usepackage{float}
\usepackage{caption}
\usepackage{cuted}
\usepackage{hyperref}

\IEEEoverridecommandlockouts               
\begin{document}

\title{\LARGE \bf
Noise Fusion-based Distillation Learning \\ 
for Anomaly Detection in Complex Industrial Environments
}

\author{Jiawen Yu$^{1}$, Jieji Ren$^{2}$, Yang Chang$^{1}$, Qiaojun Yu$^{2,3}$, Xuan Tong$^{1}$, \\
Boyang Wang$^{1}$, Yan Song$^{1}$, You Li$^{1}$, Xinji Mai$^{1}$ and Wenqiang Zhang$^{1*}$
\thanks{$^{*}$Corresponding author: Wenqiang Zhang.}%
\thanks{$^{1}$Jiawen Yu, Yang Chang, Xuan Tong, Boyang Wang, Yan Song, Xinji Mai, You Li and Wenqiang Zhang are with College of Intelligent Robotics and Advanced Manufacturing, Fudan University, Shanghai, China. {\tt\small \{jwyu23, ychang24, xtong23, bywang22, yansong24, youli22, xjmai23\}@m.fudan.edu.cn, wqzhang@fudan.edu.cn}}
\thanks{$^{2}$Jieji Ren is with Shanghai Jiao Tong University, Shanghai, China. {\tt\small jiejiren@sjtu.edu.cn}}%
\thanks{Qiaojun Yu is with $^{2}$Shanghai Jiao Tong University and $^{3}$Shanghai AI Laboratory, Shanghai, China. {\tt\small yqjllxs@sjtu.edu.cn}}%
}

\thispagestyle{empty}
\pagestyle{empty}

\maketitle

\begin{strip}
\begin{minipage}{\textwidth}\centering
\vspace{-80pt}
\includegraphics[width=.9\textwidth]{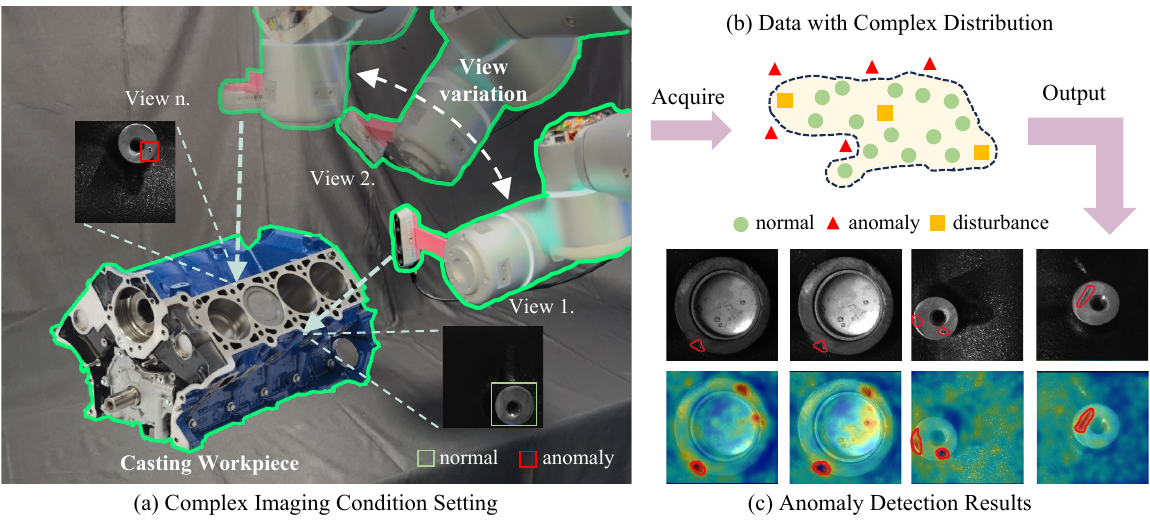}
       \captionof{figure}{(a) Illustration of our task setup. The moving robot arm introduces variations in imaging conditions like resolution, perspective and illumination. (b) Acquired image data exhibiting complex distribution, which presents challenges for models in distinguishing between normal and anomalous patterns. (c) Defects on casting workpieces in unstructured environments can be robustly highlighted by \textit{HetNet}, enhancing production efficiency.}
       \label{fig1}
\end{minipage}
\end{strip}

\begin{abstract}

Anomaly detection and localization in automated industrial manufacturing can significantly enhance production efficiency and product quality. Existing methods are capable of detecting surface defects in pre-defined or controlled imaging environments. However, accurately detecting workpiece defects in complex and unstructured industrial environments with varying views, poses and illumination remains challenging. We propose a novel anomaly detection and localization method specifically designed to handle inputs with perturbative patterns. Our approach introduces a new framework based on a collaborative distillation heterogeneous teacher network \textit{(HetNet)}, an adaptive local-global feature fusion module, and a local multivariate Gaussian noise generation module. \textit{HetNet} can learn to model the complex feature distribution of normal patterns using limited information about local disruptive changes. We conducted extensive experiments on mainstream benchmarks. \textit{HetNet} demonstrates superior performance with approximately 10\% improvement across all evaluation metrics on MSC-AD under industrial conditions, while achieving state-of-the-art results on other datasets, validating its resilience to environmental fluctuations and its capability to enhance the reliability of industrial anomaly detection systems across diverse scenarios. Tests in real-world environments further confirm that \textit{HetNet} can be effectively integrated into production lines to achieve robust and real-time anomaly detection. Codes, images and videos are published on the project website at: \href{https://zihuatanejoyu.github.io/HetNet/}{https://zihuatanejoyu.github.io/HetNet/}

\end{abstract}

\section{INTRODUCTION}

Defect detection serves as a cornerstone in manufacturing quality assurance, substantially improving production efficiency and product reliability \cite{TIM1, tim4}. This capability is particularly essential for autonomous industrial systems that identify defects without human intervention \cite{TIM2}. The detection of surface anomalies on metal components—such as micro-cracks, porosity, or deformations—remains critical as these defects can compromise structural integrity and functional performance across automotive, aerospace, and precision machinery sectors \cite{TIM3, TIM5}. The unsupervised paradigm in anomaly detection holds substantial industrial value \cite{bergmann2019mvtec}, as it circumvents the inherent challenges of limited defect samples and class imbalance, enabling more robust quality assurance in automated manufacturing processes.

With the evolution of flexible manufacturing systems, traditional fixed-position inspection stations have become increasingly inadequate. This manufacturing paradigm shift has given rise to robot-based inspection systems that can adapt to varying product geometries and production layouts. However, robot-based inspection introduces significant challenges that current anomaly detection methods struggle to address. Contemporary approaches \cite{bergmann2018ssimr1, golan2018deepr2, gong2019memorizingr3, park2020learningr4, salehi2021MKDr5, wang2021studentr6, wan2022tiiadr7}, while showing promising results in controlled environments \cite{bergmann2019mvtec, wang2024realiad}, exhibit diminished performance in real-world production settings\cite{mscad}. These methods face critical limitations when confronted with uncontrolled illumination conditions, perspective variations, and magnification differences inherent to robotic inspection systems as illustrated in Fig. \ref{fig1}. Such environmental variables introduce significant intra-class variability and generate non-defective visual patterns that are frequently misclassified as anomalies, including specular reflections, motion blur, and inconsistent textural representations.

To address these fundamental challenges, we propose HetNet, a novel framework that models the complex manifold of normal patterns in unstructured inspection scenarios. Our approach is grounded in the principle that multi-modal feature extraction mechanisms can more effectively characterize the probabilistic distribution of normal variations—essential for robust anomaly detection in complex environments. By integrating CNN architectures (optimized for hierarchical local feature extraction) with Transformer networks (specialized in modeling long-range dependencies), we simultaneously capture heterogeneous features and local-global contextual information. This dual-representation approach effectively addresses the inherent trade-off between localization sensitivity and contextual reasoning. Our adaptive local-global feature fusion module employs a dynamic attention mechanism that optimizes information flow between complementary feature spaces, while our collaborative distillation framework enforces a joint optimization objective encompassing both reconstruction and denoising tasks. This approach guides the student network toward learning a more generalizable latent representation. Additionally, our local multivariate Gaussian noise generation module introduces perturbations in the feature space, effectively expanding the decision boundary to approximate natural variation limits of normal patterns.

Our main contributions are:
\begin{itemize}
    \item A novel heterogeneous framework (HetNet) that leverages complementary feature extraction mechanisms to model complex normal variations in industrial environments with limited training data.
    
    \item An adaptive local-global feature (ALGF) fusion module with dynamic attention mechanisms bridging local sensitivity and global context awareness, and a local multivariate Gaussian noise (LMGN) generator that introduces structured perturbations approximating the natural variation boundaries of normal patterns.
    
    \item Extensive experiments on industrial benchmarks and real-world settings demonstrate robust performance improvements over existing approaches, particularly under varying environmental conditions.
\end{itemize}

\section{Related Work}

This section briefly reviews anomaly detection methods and datasets. Various datasets used in the field of anomaly detection highlight different settings, data objects, and scales. These datasets have been leveraged in numerous approaches that address the task from different perspectives, which can be broadly categorized into embedding-based and reconstruction-based approaches.

\subsection{Anomaly Detection Methods}
Embedding-based methods rely on large-scale pretrained networks \cite{krizhevsky2012pretrain} to extract features, constructing distribution models of normal-class pretrained features \cite{cohen2020spade, roth2022patchcore, defard2021padim, bae2023pni, gudovskiy2022cflow, kim2023sanflow} and identifying anomalies based on feature distance from the normal distribution. Other approaches use deep one-class classification models for single-class and novelty detection \cite{li2021cutpaste, schluter2022nsa, liu2023simplenet, lee2022cfa}. Methods based on knowledge distillation \cite{hinton2015kdistilling} employ a student network to learn from a pretrained teacher network on normal data, with output discrepancies used to localize defects \cite{bergmann2020uninformedStudents, wang2021STPM, deng2022rd4ad, gu2023memkd, zhang2023destseg}. Although embedding-based methods are computationally efficient due to the use of pre-trained networks, they often exhibit lower detection accuracy.

Reconstruction-based methods utilize generative models to reconstruct images or pretrained features, aiming to detect anomalies through reconstruction errors \cite{ lu2023removing, zhang2023diffad, zavrtanik2021draem, you2022adtr, you2022uniad, guo2023template, yao2023fod, zhao2023omnial}. Advanced models like GANs, Transformers, and Diffusion models \cite{goodfellow2020gan, vaswani2017transformer, ho2020diffusion} have been used to enhance reconstruction quality. The reverse distillation paradigm \cite{deng2022rd4ad} introduces an encoder-decoder approach to increase representation discrepancies, further improved by subsequent works \cite{tien2023rd++, guo2024recontrast, zhang2024invad}. While these methods can achieve better localization accuracy, they depend heavily on the quality and quantity of training data, often leading to increased false positives when data is insufficient.

\begin{figure*}[!t]
\centerline{\includegraphics[width=2\columnwidth]{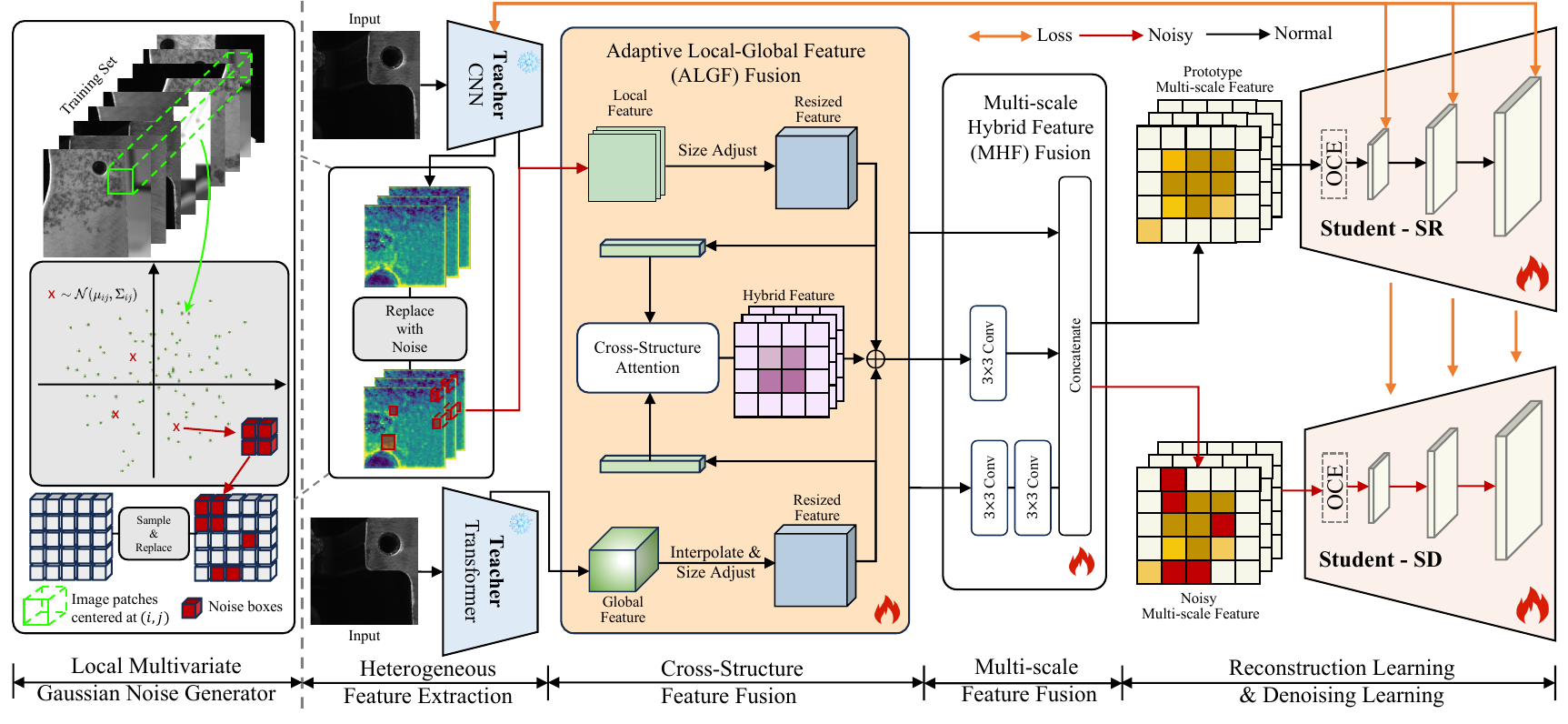}}
\caption{The overall framework of HetNet. The heterogeneous teacher encoders initially extract pre-trained features, after which the Local Multivariate Gaussian Noise (LMGN) generator introduces noise before the features are passed to the ALGF module. Subsequently, cross-structural and multi-scale features are fused through the ALGF and MHF modules. The prototype and noisy multi-scale features are then sent to the student decoder for reconstruction and denoising training.}
\label{fig3}
\end{figure*}

\subsection{Anomaly Detection Datasets}
Image anomaly detection datasets fall into two categories: synthetic and real-world objects. Synthetic datasets \cite{wieler2007dagm, tao2018cplid, bao2023miad} are generated using computer-aided design software, which, while cost-effective and scalable, fail to capture natural factors like real camera distortions. Real-world datasets \cite{song2013neu, han2023ssgd, mishra2021vtadl, bergmann2019mvtec, bergmann2022mvtec-loco, zou2022visa, bai2023vision, wang2024realiad, zhang2024pku-goodsad} focus on industrial parts, capturing images under controlled conditions, thus offering high-quality data. Given the importance of metal surfaces in industry, datasets specifically targeting metal defect detection \cite{yang2021steeltube, tsang2016fabric, tabernik2020kolektor, bovzivc2021kolektor2, huang2020mtd} are crucial. However, these datasets are limited by their simplified lab environments, unlike the complexities found in real factories.

To bridge the gap between defect detection techniques and industrial environments, the MSC-AD dataset \cite{mscad} was created, simulating multi-scene anomaly detection in real-world settings. With controlled light intensity and resolution settings, MSC-AD provides a realistic and challenging benchmark for evaluating industrial anomaly detection methods. Despite the critical role of metal surface defect detection in industrial production, it has not yet been thoroughly explored, highlighting the need for in-depth investigation.

\section{Method}

\subsection{Framework Overview}

To address the challenge of avoiding misleading disturbances when detecting anomalies in complex industrial environments, the model should become more robust to highly variable inputs. Therefore, we propose the HetNet architecture, which learns complex normal pattern feature distributions containing local disruptive variations.

HetNet comprises four main components: a heterogeneous teacher network, a cross-structure feature fusion module, a multi-scale feature fusion module, and a collaborative student network. The image is input into two heterogeneous encoders to obtain local and global representations. The ALGF module fuses these features, facilitating effective interaction between global and local characteristics. The MHF module then combines features from different layers, passing them to the student decoder. The collaborative student more accurately model the normal patterns by denoising and reconstruction.

The following sections describe the heterogeneous teacher encoder, the hybrid feature fusion strategy, and the collaborative learning with local noise. The cross-structure and multi-scale feature fusion modules will be discussed together in the hybrid feature fusion strategy.

\subsection{Heterogeneous Teacher Encoder}
Existing reverse distillation methods often result in false positives in unpredictable environments, primarily due to their reliance on convolutional neural networks (CNNs) with limited receptive fields that overemphasize local features and are sensitive to abrupt changes. To mitigate these issues, we employ CNNs and Transformer networks as teacher networks to capture both local details and long-range dependencies, making the model robust to local perturbations.

In HetNet, the Transformer teacher network \(T_{\text{Trans}}\) extracts global feature information, while the CNN teacher network \(T_{\text{CNN}}\) focuses on local feature representation, crucial for detecting small anomalies. Both networks are pretrained on ImageNet\cite{deng2009imagenet} with their parameters frozen during training.

Mathematically, let \(I\) represent the original input image, and the corresponding layer features in our heterogeneous teacher encoders are represented as \(\{f_{Global}^k = T_{\text{Trans}}^k(I), f_{Local}^k = T_{\text{CNN}}^k(I)\}\), where \(T_{\text{Trans}}^k\) and \(T_{\text{CNN}}^k\) denote the \(k^{\text{th}}\) encoding layers in the Transformer and CNN, respectively. Here, \(f_{Global}^k \in \mathbb{R}^{C_{\text{Trans}}^k \times H_{\text{Trans}}^k \times W_{\text{Trans}}^k}\) and \(f_{Local}^k \in \mathbb{R}^{C_{\text{CNN}}^k \times H_{\text{CNN}}^k \times W_{\text{CNN}}^k}\), where \(C\), \(H\), and \(W\) represent the number of channels, height, and width of the feature maps. For the subsequent knowledge distillation, we need to fuse these feature maps from different sources. First, the smaller-sized Transformer feature map is resized via interpolation to match the dimensions of the corresponding CNN feature map, resulting in the new \(\tilde{f}_{Global}^k \in \mathbb{R}^{C_{\text{Trans}}^k \times H_{\text{CNN}}^k \times W_{\text{CNN}}^k}\).

\subsection{Hybrid Feature Fusion Strategy}

Different architectures capture distinct feature properties: CNNs focus on local details, while Transformers excel at modeling long-range dependencies. To effectively integrate these complementary representations, we propose a dynamic hybrid feature fusion strategy.

Our strategy comprises two modules: the Adaptive Local-Global Feature (ALGF) fusion module and the Multi-Scale Hybrid Feature (MHF) fusion module. ALGF operates at each layer, performing local-global cross-structural attention fusion to generate hybrid features. These hybrid features are then further fused through MHF, resulting in multi-scale hybrid features.

Within a single ALGF block, we first align two teacher network feature channel numbers with convolution blocks and flatten them to get \(f_{Local}'^k, f_{Global}'^k \in \mathbb{R}^{C^k \times H^k W^k}\).

Before flattening, we concatenate the two heterogeneous features along the channel dimension to obtain \(g^k \in \mathbb{R}^{2C^k \times H^k \times W^k}\) for subsequent residual connection. Next, we use the feature \(f_{Local}'^k\) as the query and value, and \(f_{Global}'^k\) as the key to perform cross-structure attention. We then apply a convolution to increase the channel size and reshape the output:
\begin{equation}
    h^k = \text{Conv}(\text{Attention}(f_{Local}'^k, f_{Global}'^k, f_{Local}'^k)).
\end{equation}

Finally, the output feature \(h^k\) from the attention module is added to the concatenated feature \(g^k\), and a convolution is applied to further fuse these features, resulting in the output hybrid feature:
\begin{equation}
    o^k = \text{Conv}(h^k + g^k) \in \mathbb{R}^{C^k \times H^k \times W^k}.
\end{equation}

After independently computing \(o^k\) for multiple layers, where \(k \in \{1, 2, ..., n\}\), we further apply the MHF module to fuse multi-scale hybrid features, making the features passed into the bottleneck structure more compact and enriched. Specifically, similar to the approach in\cite{deng2022rd4ad}, shallow features are downsampled using one or more 3×3 convolutional layers with a stride of 2, followed by batch normalization and ReLU activation. The resized features from each layer are then concatenated. The output of the hybrid feature fusion is the prototype multi-scale feature \(o\).

\begin{table*}[h]
\caption{Anomaly detection and localization results in terms of Image-AUROC / Pixel-AUROC / Pixel-AUPRO on the MSC AD dataset. Best results in \textbf{bold}, N indicates the unavailable results.}
\label{table1}
\centering
\renewcommand\arraystretch{0.9}
\resizebox{2\columnwidth}{!}{%
\begin{tabular}{ccccccc}
\toprule
\multicolumn{2}{c}{Condition}              & DRAEM\cite{zavrtanik2021draem} & CFA\cite{lee2022cfa}         & CFLOW\cite{gudovskiy2022cflow}  & RD4AD\cite{deng2022rd4ad}  & HetNet(Ours)           \\ \cmidrule{1-7}
\multicolumn{1}{c}{\multirow{4}{*}{Resolution}}   & 150x150 & 54.40 / 77.39 / N    & 65.34 / N / 85.65          & 52.33 / 87.03 / 60.80 & 42.62 / 86.24 / 69.86 & \textbf{83.31 / 97.42 / 93.34} \\
\multicolumn{1}{c}{}             & 200x200 & 57.08 / 80.17 / N    & 73.68 / N / 88.62 & 60.71 / 88.50 / 62.54 & 48.88 / 90.01 / 76.81 & \textbf{88.94 / 98.40 / 96.01} \\
\multicolumn{1}{c}{}             & 300x300 & 57.07 / 82.21 / N    & 66.62 / N / 88.01          & 53.00 / 86.84 / 65.94 & 42.42 / 89.96 / 77.37 & \textbf{85.57 / 98.09 / 94.95} \\
\multicolumn{1}{c}{}                      & 600x600 & 47.01 / 87.10 / N    & 63.78 / N / 87.49          & 60.35 / 90.56 / 66.49 & 41.30 / 91.21 / 78.10 & \textbf{85.41 / 97.98 / 94.55} \\ \cmidrule{1-7}
\multicolumn{1}{c}{\multirow{3}{*}{Illumination}} & high    & 50.86 / 81.81 / N    & 67.12 / N / 87.10          & 52.73 / 87.44 / 63.30 & 42.53 / 89.70 / 76.75 & \textbf{83.18 / 97.68 / 94.77} \\
\multicolumn{1}{c}{}                      & mid     & 56.49 / 81.67 / N    & 68.75 / N / 87.53          & 56.85 / 88.51 / 63.94 & 43.54 / 89.55 / 76.16 & \textbf{86.62 / 98.17 / 94.89} \\
\multicolumn{1}{c}{}                      & low     & 54.32 / 81.68 / N    & 66.19 / N / 87.78          & 60.21 / 88.75 / 64.33 & 45.35 / 88.80 / 73.69 & \textbf{87.62 / 98.05 / 94.48} \\ \cmidrule{1-7}
\multicolumn{1}{c}{\multirow{6}{*}{Surface}}      & back    & 20.11 / 83.10 / N    & 51.62 / N / 88.97          & 32.01 / 86.71 / 49.28 & 38.33 / 91.04 / 78.40 & \textbf{73.16 / 96.55 / 93.19} \\
\multicolumn{1}{c}{}                      & down    & 69.80 / 89.84 / N    & 72.94 / N / 92.40          & 65.44 / 89.11 / 72.53 & 41.90 / 90.44 / 74.60 & \textbf{87.30 / 98.30 / 95.37} \\
\multicolumn{1}{c}{}                      & front   & 88.69 / 91.29 / N    & 88.88 / N / 96.55          & 69.90 / 9..87 / 76.55 & 54.80 / 92.58 / 82.36 & \textbf{93.30 / 99.36 / 97.39} \\
\multicolumn{1}{c}{}                      & left    & 44.25 / 65.89 / N    & 71.30 / N / 89.71          & 45.53 / 87.10 / 54.47 & 37.75 / 83.98 / 72.23 & \textbf{89.89 / 98.11 / 92.26} \\
\multicolumn{1}{c}{}                      & right   & 42.91 / 72.38 / N    & 54.63 / N / 88.60          & 43.53 / 83.97 / 58.64 & 36.76 / 87.08 / 68.02 & \textbf{78.20 / 97.47 / 95.96} \\
\multicolumn{1}{c}{}                      & up      & 57.57 / 87.82 / N    & 64.75 / N / 68.58          & 83.16 / 88.65 / 72.86 & 53.30 / 91.01 / 77.60 & \textbf{93.00 / 98.03 / 94.11} \\ \cmidrule{1-7}
\multicolumn{2}{c}{\textit{Total Average}}                   & 53.89 / 81.72 / N    & 67.35 / N / 87.47          & 56.60 / 88.23 / 63.88 & 43.81 / 89.35 / 75.53 & \textbf{85.81 / 97.97 / 94.71} \\ 
 \bottomrule
\end{tabular}%
}
\vspace{-0.2cm}
\end{table*}

\begin{figure*}[!t]
\centerline{\includegraphics[width=2\columnwidth]{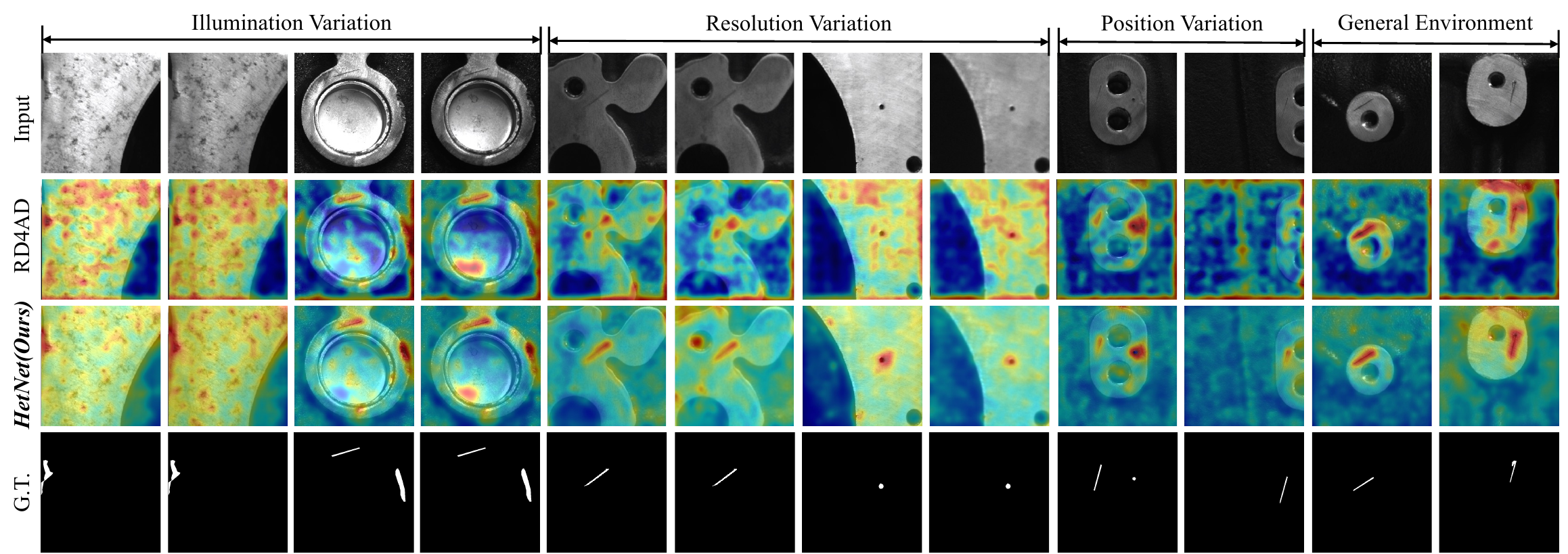}}
\caption{Qualitative results on the MSC-AD dataset\cite{mscad}, demonstrating the efficacy of HetNet. From left to right, we present anomaly detection results for samples under varying illumination, resolution, and positioning conditions. Predictions made by HetNet robustly adapt to input disturbances and exhibit more stable performance across diverse environmental conditions. The lower image resolution or brightness only slightly affect the model accuracy.}
\label{fig4}
\vspace{-0.2cm}
\end{figure*}

\subsection{Collaborative Learning with Local Noise}
Current reverse distillation methods rely on information-rich training data, but their generalization ability is limited when disturbance patterns are scarce, leading to reconstruction errors and false positives. To address this, we introduce collaborative distillation, where the student network performs both reconstruction and denoising tasks simultaneously. This improves feature quality and enhances robustness to normal disturbances. The noise for denoising is generated by a Local Multivariate Gaussian Noise (LMGN) generator.

Let the prototype multi-scale feature \(o\) be input into the student network \(SR^k\) for direct reconstruction, and the noisy multi-scale feature \(o'\) be input into the student network \(SD^k\) for denoising. We will discuss the process of constructing noisy feature \(o'\) later. These two student networks share parameters and each contains an OCE bottleneck module as in \cite{deng2022rd4ad}, as well as a decoder with the same structure as the teacher CNN but reversed. The activations of the student network are denoted as \(\{f_{SR}^k = SR^k(o), f_{SD}^k = SD^k(o')\}\), collectively referred to as \(f_S^k\). Following standard practice\cite{deng2022rd4ad,hinton2015kdistilling}, we compute the vector-wise cosine similarity loss along the channel axis between the CNN teacher network \(f_{Local}^k\) and the student network \(f_S^k\), obtaining a 2-D anomaly map \(M^k \in \mathbb{R}^{H_{\text{CNN}}^k \times W_{\text{CNN}}^k}\). The objective functions are computed as:
\begin{equation}
    L_{KD/Recon} = \sum_{k=1}^K \{\frac{1}{H_kW_k}\sum^{H_k}_{h=1}\sum^{W_k}_{w=1}M^k_{SR/SD}(h,w)\},
\end{equation}

where \(k\) indicates the number of feature layers used in the experiment. The overall loss is defined as:
\begin{equation}
    L_{Total} = L_{KD} + \alpha L_{Recon},
\end{equation}
where \(\alpha\) is a regularization parameter.

To construct the noisy multi-scale feature \(o'\) mentioned above, we design a new LMGN generator to introduce normal disturbance information into the training features. Inspired by \cite{defard2021padim}, we use a multivariate Gaussian distribution to describe the probabilistic representation of each position for the normal class and sample noise from this normal distribution to simulate complex normal variation patterns. We first use the CNN teacher network \(T_{\text{CNN}}^k\) to extract pretrained features from all images in the training set and take the feature vectors at the same center position \((i, j)\) for all images, describing the normal feature distribution at different center positions \((i, j)\). At each center position \((i, j)\), we calculate the mean \(\mu_{ij}\) and covariance \(\Sigma_{ij}\) following \cite{defard2021padim}.

During training, we sample a noise of random size \(\xi \sim \mathcal{N}(\mu_{ij}, \Sigma_{ij})\) from the above distribution and add it into features extracted from the teacher network \(f_{Local}^k\). This process introduces local scale noise at different center positions of the feature map to obtain the final noisy feature, which is then fed into the subsequent ALGF module. The multivariate Gaussian distribution describes the distribution of normal features in the training set, and the noise sampled from this distribution is related to the normal features but has some randomness, enabling the student network to be more robust to slight variations that may exist in the input.

\section{Experiments}

We validated our method through extensive experiments on the MSC-AD dataset\cite{mscad}, demonstrating its effectiveness in complex environments. Additional tests on the MVTec-AD, VisA and MPDD datasets\cite{bergmann2019mvtec, zou2022visa, mpdd} confirmed its generalization ability and robustness. Real-world environment tests further substantiated its practical application.

\subsection{Experiments on the MSC-AD Dataset}
\textbf{1) Training Details:} %

Similar to\cite{deng2022rd4ad}, we use WideResNet50\cite{zagoruyko2016wideresnet} as the CNN encoder. Given the efficiency requirements in industrial defect detection, we selected the lightweight Swin-T-22k\cite{liu2021swinT} as the Transformer encoder, resizing images to 256 resolution and utilizing the intermediate layers with \(k = \{1, 2, 3\}\). We employed the Adam Optimizer\cite{kingma2014adam} with betas=(0.5, 0.999), setting the learning rate to 0.005, and the regularization parameter of \(L_{\text{Recon}}\) to 0.1. We trained for 200 epochs with a batch size of 16.

\textbf{2) Inference Details:}
During the inference stage, we maintain the overall framework but discard the LMGN generator. We extract the corresponding layer representation pairs from the CNN encoder and student decoder, compute the differences between the pairs at different layers, and apply bilinear upsampling to restore the difference maps to the input image size. Finally, we aggregate the anomaly maps from all layers to generate the final anomaly localization score map. For anomaly detection, we take the maximum value from the anomaly score map as the image-level anomaly score.

\textbf{3) Experiment Results and Discussions:}
We use the area under the receiver operating characteristic (AUROC) for evaluating image-level anomaly detection and pixel-level anomaly localization. To address AUROC’s bias toward large anomalies \cite{bergmann2020uninformedStudents}, we also utilize the per-region-overlap (PRO) metric for a fairer evaluation across different anomaly sizes.

Table.\ref{table1} presents the average results across different imaging conditions, calculated following \cite{mscad}. HetNet outperforms SOTA methods in all categories on the MSC AD dataset, improving image-level AUROC by 18.46\% over the second-best method \cite{lee2022cfa} and nearly 40\% over others. This indicates more accurate anomaly detection, likely due to the collaborative student network’s ability to delineate normal and abnormal feature boundaries on complex industrial workpieces, without being influenced by environmental variations.

AUROC favors larger anomaly regions, while AUPRO provides a more balanced evaluation of anomalies of different sizes \cite{bergmann2020uninformedStudents}. As shown in Table.\ref{table1}, HetNet achieves superior average pixel-level AUPRO results compared to SOTA methods, indicating its effectiveness in detecting and localizing small anomalies. Qualitative results in Fig.\ref{fig4} further demonstrate its capability in detecting metal surface defects in complex industrial environments.

\subsection{Ablation Study}

We conducted ablation experiments on the MSC-AD dataset to exhibit the effectiveness of each module in HetNet. Experiments were conducted on the difficult subset with 150 resolution classes under high and low illumination, as this subset is the most challenging and representative of real industrial environments. Results on additional datasets further evaluate our method’s generalization capability.

\begin{table}[h]
\caption{Ablation study results of different modules.}
\label{table3}
\centering
\renewcommand\arraystretch{0.9}
\resizebox{\columnwidth}{!}{%

\begin{tabular}{cccccc}
\toprule
TE    & ALGF & LMGN & I-AUROC & P-AUROC & P-AUPRO \\
\cmidrule{1-6}
CNN   & \ding{55}    & \ding{55}    & 42.02   & 86.22   & 69.59   \\
Trans & \ding{55}    & \ding{55}    & 49.78   & 43.80   & 58.62   \\
HTE   & \ding{55}    & \ding{55}    & 80.98   & 96.93   & 91.76   \\
HTE   & \checkmark    & \ding{55}    & 83.46   & 97.18   & 92.30   \\
HTE   & \ding{55}    & \checkmark    & 84.02   & 97.23   & 92.36   \\
HTE   & \checkmark    & \checkmark    & \textbf{84.20}   & \textbf{97.30}   & \textbf{92.85}   \\
\bottomrule
\end{tabular}
}
\vspace{-0.2cm}
\end{table}

\textbf{1) Teacher Network Structure:}
The first three rows of Table.\ref{table3} show the importance of the heterogeneous teacher encoder (HTE). Using only a standalone CNN or Transformer network results in lower performance than the heterogeneous teacher encoder. This validates the combination of Transformer and CNN in HTE.

\textbf{2) Module Ablation Design:}
The last four rows of Table.\ref{table3} present results for different architecture configurations. (1) HetNet achieves strong performance even without the feature fusion and noise modules, indicating the effectiveness of HTE. (2) Performance improves with the addition of the ALGF and LMGN modules. (3) Including both modules yields the best results in anomaly detection and localization.

\begin{table}[h]
\caption{Ablation study results of different noise types.}
\label{table4}
\centering
\renewcommand\arraystretch{0.9}
\fontsize{9}{10}\selectfont{\begin{tabular}{cccc}
\toprule
Noise    & I-AUROC      & P-AUROC & P-AUPRO \\
\cmidrule{1-4}
\ding{55}    & 83.46        & 97.18   & 92.30   \\
randn    & 83.89        & 97.27   & 92.76   \\
mGds(Ours)    & \textbf{84.20}        & \textbf{97.30}   & \textbf{92.85}   \\

\bottomrule
\end{tabular}
}
\end{table}

\textbf{3) Choice of Noise Type:}
Table.\ref{table4} shows the impact of different noise types. Results show that introducing standard normal distribution noise (randn) slightly improves performance, but sampling noise from a multivariate Gaussian distribution (mGds) yields optimal performance, enhancing model robustness.

\begin{figure}[!t]
\centerline{\includegraphics[width=\columnwidth]{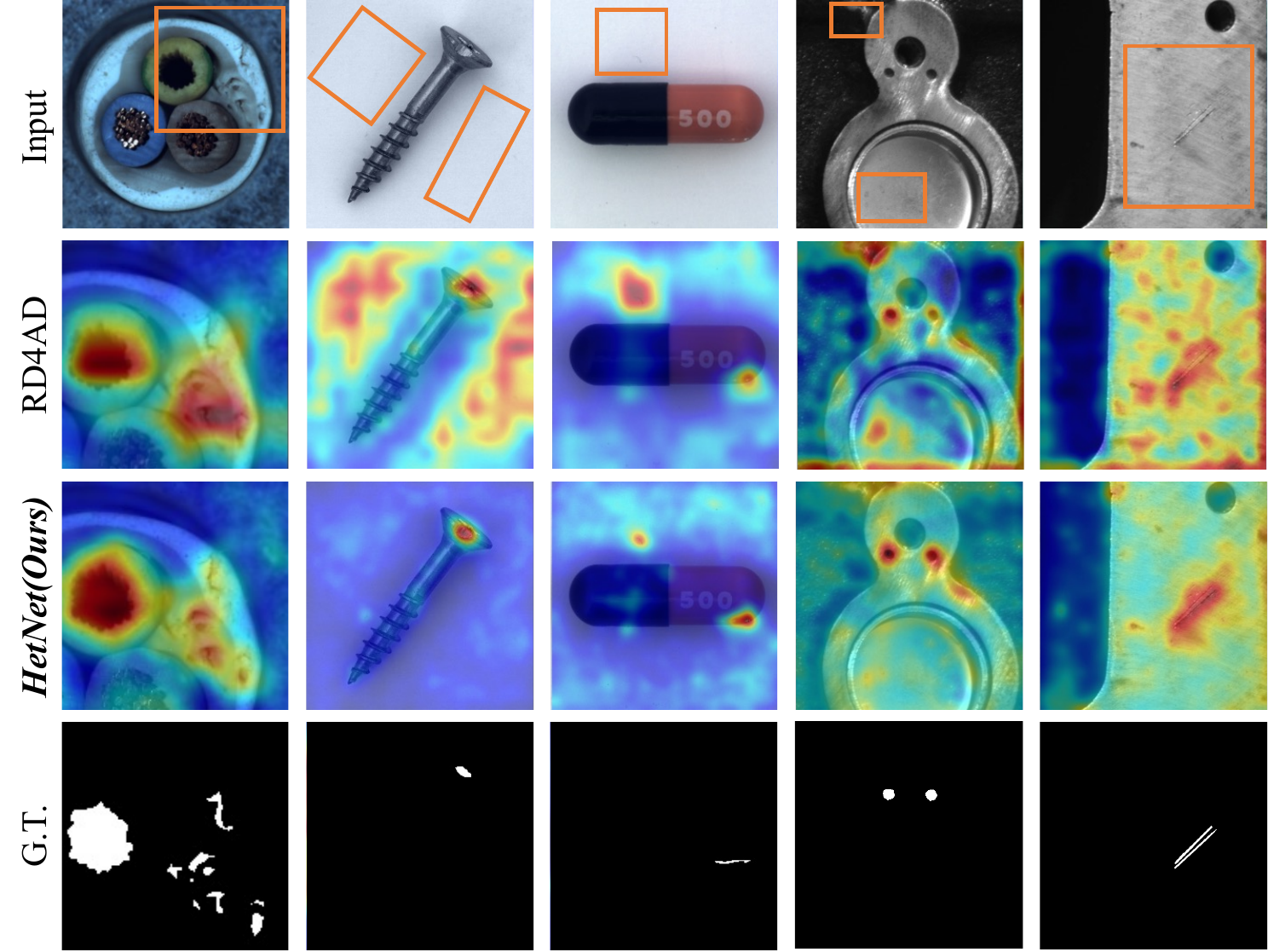}}
\caption{Qualitative comparisons demonstrating HetNet's effectiveness in reducing false positives on normal disruptive patterns. The highlighted regions in the first row show areas where baseline methods produce significant false positives or ambiguous predictions. First three columns are from MVTec-AD\cite{bergmann2019mvtec}, while last two columns are from MSC-AD\cite{mscad}.}
\label{fig5}
\vspace{-0.2cm}
\end{figure}

\textbf{4) Results on Additional Datasets:}
As demonstrated in Table~\ref{table2}, the proposed HetNet achieves state-of-the-art performance across three diverse public benchmarks. On the MVTec AD dataset, our approach attains the highest image-level AUC score of 99.19\% and pixel-level PRO score of 94.63\%, which can be attributed to the heterogeneous feature encoding mechanism that effectively captures both global semantic inconsistencies and local structural anomalies. For the more challenging VisA dataset, HetNet significantly outperforms existing methods with an I-AU of 96.19\% and P-PRO of 94.06\%. This substantial enhancement stems from our multi-scale feature fusion strategy that adaptively balances high-level semantic information with low-level textural details, particularly beneficial for the diverse anomaly types present in VisA. On the MPDD dataset, which features complex surface defects under varying illumination conditions, HetNet demonstrates competitive performance with the highest P-PRO score of 95.43\% while maintaining strong results across other evaluation metrics. The consistent superiority across these datasets with distinct characteristics (controlled lighting in MVTec, complex backgrounds in VisA, and illumination variations in MPDD) empirically validates the theoretical foundation of our approach: that heterogeneous feature interaction effectively bridges the representational gap between normal and anomalous patterns across different industrial inspection scenarios, leading to improved generalization capabilities without requiring domain-specific adaptations. More qualitative results can be found in Fig.\ref{fig5}

\begin{table*}[]
\caption{Results on Additional datasets.}
\label{table2}
\centering
\resizebox{1.5\columnwidth}{!}{%
\begin{tabular}{cccccccccc}
\toprule
\textbf{}  \multirow{2}{*}{Method}   & \multicolumn{3}{c}{MVTec AD}                     & \multicolumn{3}{c}{VisA}                         & \multicolumn{3}{c}{MPDD}                       \\ 
      & I-AU        & P-AU        & P-PRO        & I-AU        & P-AU        & P-PRO        & I-AU       & P-AU       & P-PRO        \\ \hline
DRAEM{\cite{zavrtanik2021draem}}  & 98.00          & 97.30          & N              & 92.4           & 92.0           & 78.7           & 94.3          & 90.7          & 78.0           \\
CFA\cite{lee2022cfa}    & 99.17          & 98.15          & 94.44          & 92.0           & 84.3           & 55.1           & 92.3          & 94.8          & 83.2           \\
CFLOW\cite{gudovskiy2022cflow}  & 98.26          & \textbf{98.62} & 94.60          & 87.77 & 98.04          & 85.29          & 87.11         & 97.42         & 88.56          \\
RD4AD\cite{deng2022rd4ad}  & 98.46          & 97.81          & 93.93          & 92.4           & 96.1           & 93.0           & \textbf{95.3} & \textbf{98.7} & 95.2           \\
HetNet(Ours) & \textbf{99.19} & 98.04          & \textbf{94.63} & \textbf{96.19} & \textbf{98.78} & \textbf{94.06} & 92.76         & 98.57         & \textbf{95.43} \\ 
\bottomrule
\end{tabular}%
}
\end{table*}

\subsection{Real Environment Experiments}
\begin{figure}[!t]
\centerline{\includegraphics[width=\columnwidth]{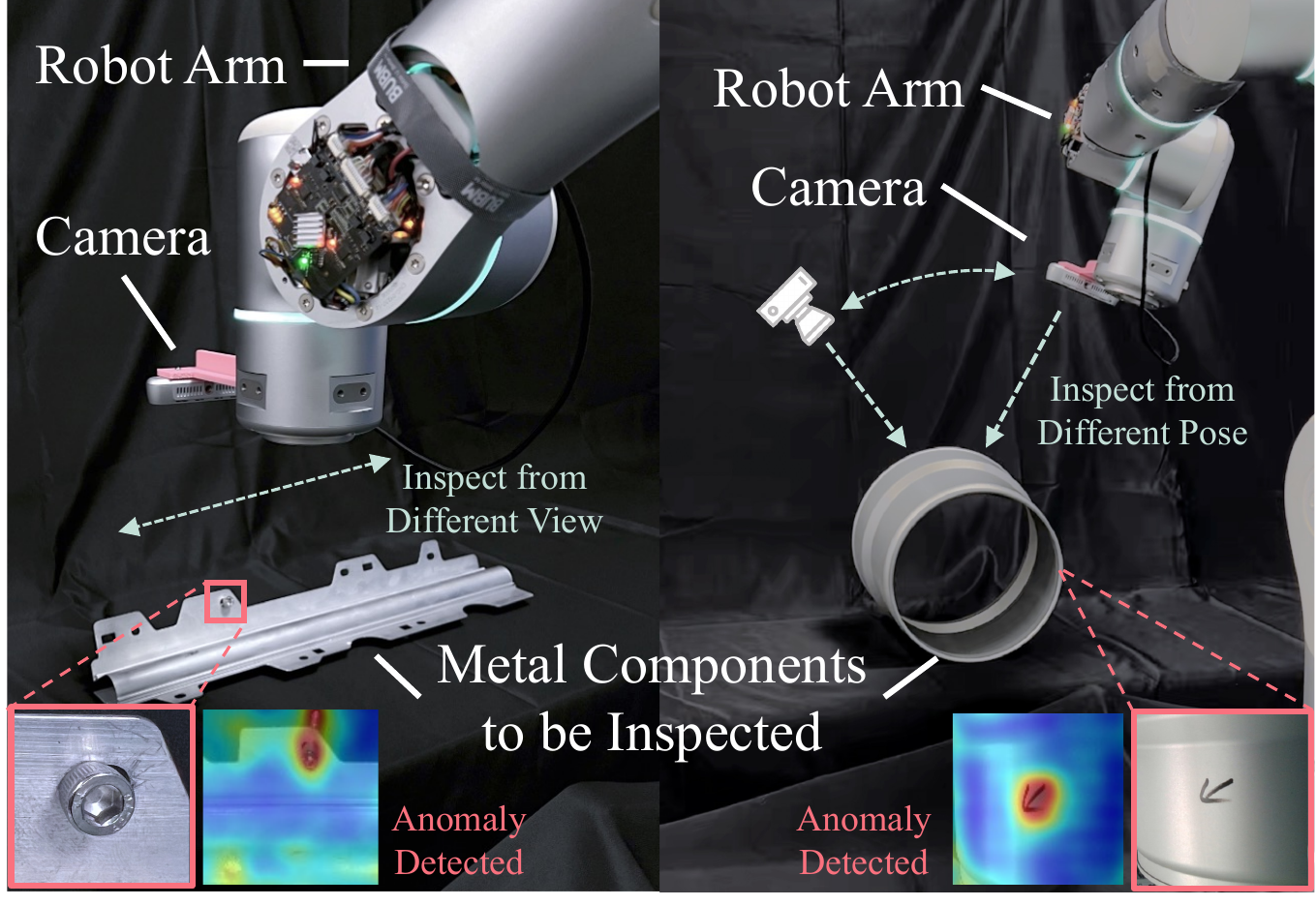}}
\caption{The real-world environment setting.}
\label{fig_real}
\end{figure}

As shown in Fig.~\ref{fig_real}, after validating HetNet's effectiveness on public datasets, we deployed it in a real-world industrial inspection setting to detect surface defects on precision metal components. The detection platform integrates Flexiv Rizon 7-Axis Robot Arm equipped with Intel RealSense D415 camera, connected to a dedicated computing system. During the inspection process, the robotic arm strategically maneuvers the camera along different positions and angles of the metal components, enabling our system to capture comprehensive multi-view images. This dynamic capture approach inherently introduces one of the key challenges in industrial inspection—the varying appearance of surfaces under different viewing conditions—which aligns with our method's robust feature extraction capabilities demonstrated on the benchmark datasets. The controlled movement of the camera also enables comprehensive detection of defects that would be imperceptible from a single viewpoint. Our HetNet algorithm operates on an Intel i7-8565u processor and a GeForce RTX 3080, utilizing approximately 1.5 GB of VRAM. The time cost for detecting a single image is only 0.27 s to 0.33 s, demonstrating high efficiency and meeting the speed requirements of industrial automated production lines. These practical deployment results validate HetNet's potential for real-world application in high-precision manufacturing inspection scenarios where consistent performance across varying inspection conditions is critical.

\section{CONCLUSIONS}

In this work, we present HetNet, an approach to anomaly detection and localization in complex industrial environments. Our model learns the distribution of intricate normal patterns from data with local variations. We introduce three key components: a heterogeneous teacher encoder that exploits the complementary strengths of different networks, a fusion module to harmonize heterogeneous teacher-student networks, and a noise generator to introduce complex variation information, improving the model’s robustness to diverse inputs. Our method achieves state-of-the-art performance on mainstream benchmarks and proves its high practical value in real-world environment tests.


\bibliographystyle{IEEEtran}
\bibliography{reference}

\end{document}